\documentclass[journal,onecolumn]{IEEEtran}

\usepackage{cite}
\usepackage{amsmath,amssymb,amsfonts}
\usepackage{algorithmic}
\usepackage{graphicx}
\usepackage{textcomp}
\usepackage{xcolor}
\usepackage[bookmarks=false]{hyperref}

\graphicspath{ {.} }
\usepackage{subcaption} 

\begin{document}

\markboth{A Preprint}{}

\title{Unsupervised Learning of Depth and Ego-Motion from Cylindrical Panoramic Video with Applications for Virtual Reality}

\author{%
    \IEEEauthorblockN{%
    Alisha~Sharma\IEEEauthorrefmark{1}\thanks{\IEEEauthorrefmark{1}Work performed as part of a NSF Research Experiences for Undergraduates (REU) program at the University of Colorado, Colorado Springs.},
    Ryan~Nett\IEEEauthorrefmark{2},
    and~Jonathan~Ventura%
        \IEEEauthorrefmark{2}%
        \IEEEauthorrefmark{3}%
        \thanks{\IEEEauthorrefmark{3}Corresponding author email: \texttt{jventura09@calpoly.edu}}
    }\\
    \IEEEauthorblockA{\IEEEauthorrefmark{1}%
        Naval Research Laboratory,
        Washington, D.C.
    }\\
    \IEEEauthorblockA{\IEEEauthorrefmark{2}%
        California Polytechnic State University,
        San Luis Obispo, CA
    }
}

\maketitle

\begin{abstract}
We introduce a convolutional neural network model for unsupervised learning of depth and ego-motion from cylindrical panoramic video.  Panoramic depth estimation is an important technology for applications such as virtual reality, 3D modeling, and autonomous robotic navigation.  In contrast to previous approaches for applying convolutional neural networks to panoramic imagery, we use the cylindrical panoramic projection which allows for the use of the traditional CNN layers such as convolutional filters and max pooling without modification.  Our evaluation of synthetic and real data shows that unsupervised learning of depth and ego-motion on cylindrical panoramic images can produce high-quality depth maps and that an increased field-of-view improves ego-motion estimation accuracy. We create two new datasets to evaluate our approach: a synthetic dataset created using the CARLA simulator, and Headcam, a novel dataset of panoramic video collected from a helmet-mounted camera while biking in an urban setting.  We also apply our network to the problem of converting monocular panoramas to stereo panoramas.
\end{abstract}


\section{Introduction}

Understanding the structure of a 3D scene is an important problem in many fields, from autonomous vehicle navigation to free-viewpoint rendering of virtual reality (VR) content.  The ability to automatically infer scene depth in panoramic video would be especially useful for free-viewpoint rendering in a VR headset \cite{serrano2019motion}, for example.

Given a color image, the scene depth is unknown and must either be inferred from single-view or multi-view cues or acquired with a different sensor.   Single-image inference is especially interesting since most consumer content is captured from a single viewpoint without special hardware for depth estimation such as time-of-flight ranging or structured light sensors.  Unfortunately, predicting 3D structure from a single image is extremely challenging. The number of confounding factors (e.g. varied texture, lighting, occlusions, and object movement) makes it an ill-posed problem: a single image could represent many possible 3D scenes.

Early attempts at estimating scene structure from motion (also known as SfM) focused on directly analyzing factors such as the geometry and flow of the image \cite{bergen_hierarchical_1992,mur-artal_orb-slam:_2015,saxena_make3d:_2009}. However, these models were often fragile in the face of occlusions, object motion, and other inconsistent, but real-world, conditions. In the past several years, many exciting advances have been made in estimating scene structure and ego-motion---motion of the observer---using deep neural networks.

Early research relied on labeled data for training \cite{eigen_depth_2014,liu_learning_2016}. Unfortunately, labeled 3D footage is expensive to create, limiting the quantity and diversity of available training data. This limitation has triggered a promising new area of research: unsupervised SfM models, which figure out scene attributes such as depth and ego-motion without requiring labeled data. In the past few years, several unsupervised models have been proposed with comparable performance to the supervised state-of-the-art \cite{godard_unsupervised_2017,zhou_unsupervised_2017,vijayanarasimhan_sfm-net:_2017,mahjourian_unsupervised_2018,wang_learning_2018}, lowering the cost and expanding the diversity of potential training datasets.

\begin{figure}[t]
    \centering
    \includegraphics{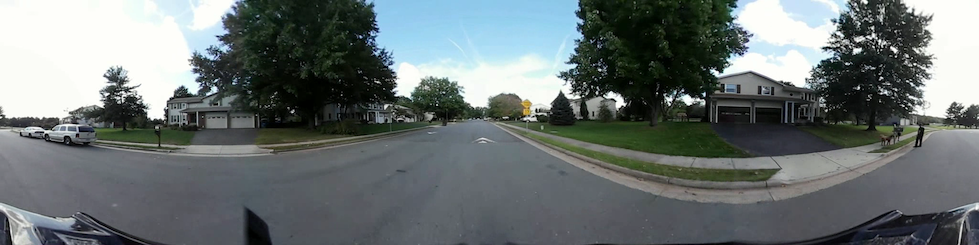}\\
    \includegraphics{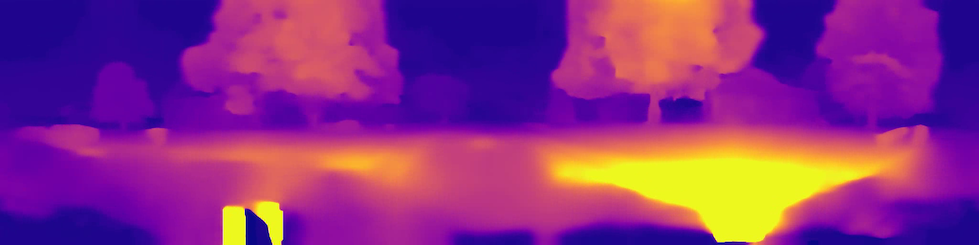}\\
    \caption{An example of predicting depth (bottom) from a single panoramic input (top) using our proposed method.  The depth is color-coded so that brighter pixels are closer.}
    \label{fig:depth_example}
\end{figure}

While much progress has been made for pinhole perspective images, researchers have only recently started to apply these deep networks to panoramic input. There are many compelling applications for computer vision with non-pinhole projection images, such as robotic vision with omnidirectional cameras.

To process panoramic imagery in a convolutional neural network (CNN), we need to choose a panoramic projection model that maps spherical coordinates to image coordinates.  The three most common options are spherical or equirectangular projection, cube map projection, and cylindrical projection.

\begin{figure}[t]
    \centering
    \includegraphics[width=0.7\linewidth]{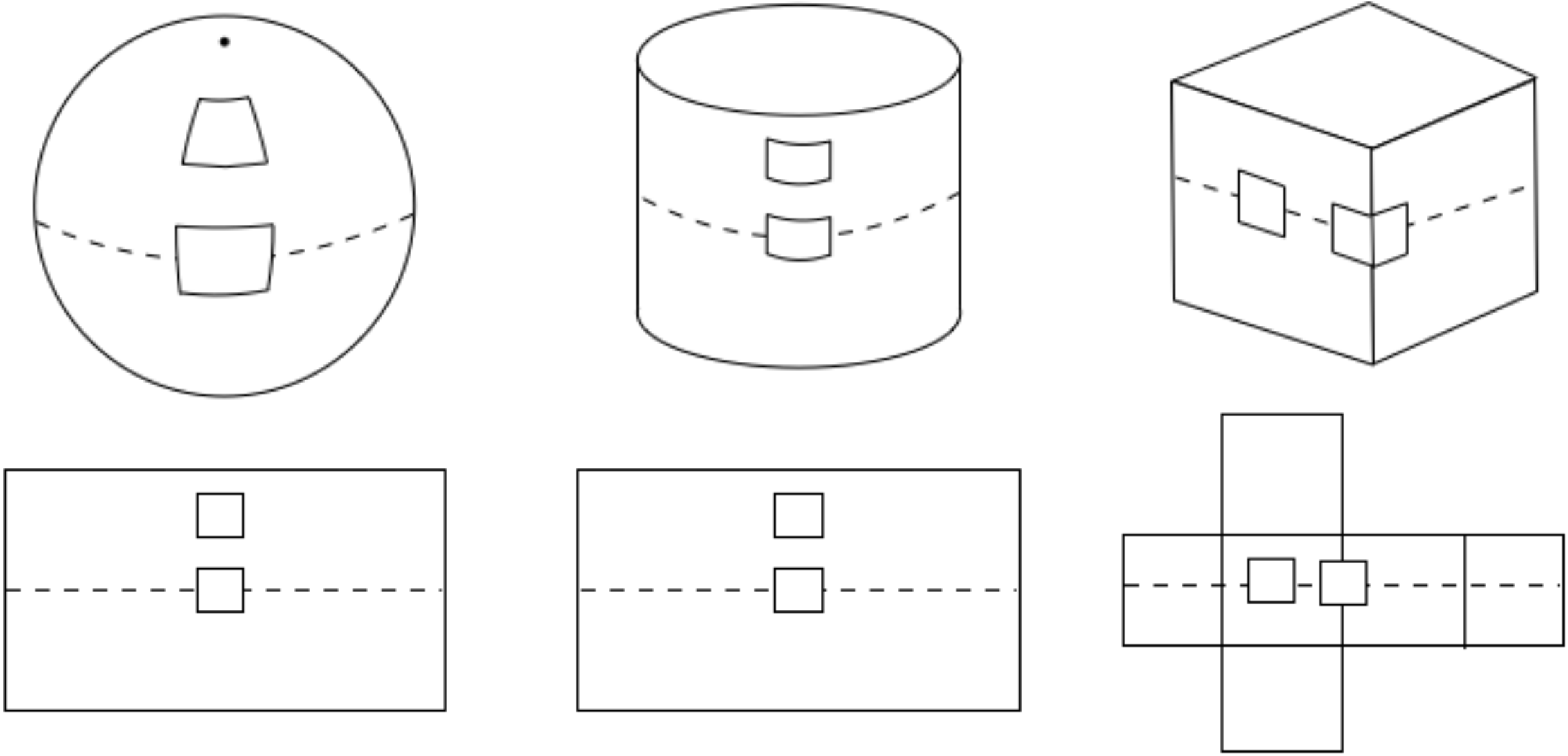}
    \caption{Comparison of convolutional filtering using equirectangular projection (left), cylindrical projection (middle), and cube map projection (right).  A square filter on the equirectangular projection (bottom left) maps to differently sized areas on the sphere (top left) \cite{cohen_spherical_2018}.  In contrast, a square filter on the cylindrical projection (bottom middle) always projects to the same area on the cylinder (top middle).  This property of cylindrical panoramas allows us to apply convolutional neural network layers to a cylindrical panorama without needing to model position-dependent effects on the receptive field, as in previous works \cite{coors_spherenet:_2018,
    tateno_distortion-aware_2018,zhao_distortion-aware_2018}.    Cube map projection (right) introduces seams and discontinuities, necessitating the use of multi-image inference and careful padding strategies \cite{Cheng_2018_CVPR,wang2018self}.}
    \label{fig:filters}
\end{figure}

Spherical projection has the advantage of representing the entire sphere in a rectangular image.  The disadvantage is the distortions at the poles caused by the projection (see Figure \ref{fig:filters}).  Because of these distortions, properly processing spherical panoramic input in a CNN requiring expensive modifications to the model layers \cite{cohen_spherical_2018,esteves_learning_2018}.

Cube map projection also represents the entire sphere and avoids distortions at the poles; however, it introduces discontinuities between the faces of the cube.  To use cube maps as input to CNNs, we need to run the model on each face separately and use a careful padding strategy and loss functions to encourage agreement between the six outputs \cite{Cheng_2018_CVPR,wang2018self}.

Cylindrical projection has been relatively less explored for CNN input.  Unlike the other projection models, cylindrical projection is continuous and avoids increasing distortion toward the poles (Figure \ref{fig:filters}).   The trade-off is that the cylinder cannot represent the entire sphere; the top and bottom are cut off.  Despite that small disadvantage, we argue that cylindrical panoramas are ideal for use in CNNs they allow for standard convolutional layers and only require a simple horizontal wrap padding.  Furthermore, in most applications, the top and bottom areas are relatively unimportant (usually consisting of sky or ceiling at the top and ground, vehicle, or camera mount at the bottom).

Despite the advantages of cylindrical projection, to our knowledge, deep networks for depth prediction have not yet been applied to cylindrical panoramic imagery.  In this paper, we address this gap by proposing and evaluating a novel unsupervised learning approach that estimates depth and ego-motion from cylindrical panoramas.  We achieve this by modifying the architecture of Zhou et al.~\cite{zhou_unsupervised_2017}, with improvements from later papers, to use cylindrical panoramic projection.

This work has four major contributions:
\begin{enumerate}
    \item We present CylindricalSfMLearner, an unsupervised model for estimating structure from motion given cylindrical panoramic input.
    \item We evaluate our method on synthetic and real data to validate our approach.
    \item We provide a new dataset of panoramic street-level videos suitable for unsupervised learning of depth and ego-motion.
    \item We demonstrate how to use the trained neural network for novel view synthesis and stereo panorama conversion from a single input panorama.
\end{enumerate}

This journal paper extends our previous conference paper \cite{aivr19} in several ways.  We evaluate our method on a larger and more complete synthetic dataset rendered using the CARLA simulator \cite{Dosovitskiy17} that allows us to more effectively evaluate our approach.  We show how our model can be used to produce stereo panoramas from monocular panoramas.  We also trained our model on higher resolution images from the Headcam dataset and show the qualitative improvement in depth maps achieved.


\section{Related Work and Background}

\subsection{Supervised Monocular Depth Prediction}

Early research focused on detecting structure from \textit{stereo}---or multi-source---imagery. Stereo SfM is much more constrained than detecting structure from \textit{monocular}---single-source---input, but the stereo input requirement limits the model's flexibility. Eigen \cite{eigen_depth_2014} proposed a different approach using deep neural networks. They presented a supervised model for estimating depth maps from monocular input images. Their model was composed of two stacks---one for coarse estimation and one for fine estimation---and joined the two predictions.

\subsection{Supervised to Unsupervised Models}

While supervised models for single-image depth prediction demonstrate excellent performance, collecting labeled footage is very expensive, increasing training cost and limiting the size and diversity of datasets. This limitation triggered some researchers to turn towards unsupervised models. Godard et al., taking inspiration from previous stereo techniques, proposed a model that was trained on unlabeled stereo footage. Their trained model outperformed the previous supervised state-of-the-art on urban scenes and performed reasonably well on unrelated datasets \cite{godard_unsupervised_2017}.

Zhou et al.~removed the constraint of needing stereo training footage \cite{zhou_unsupervised_2017}. They proposed an unsupervised model composed of jointly-trained depth and pose CNNs using a loss function tied to novel view synthesis. They found that their unsupervised model performed comparably to supervised models on the known datasets and reasonably well when tested against a completely unknown data set. Unfortunately, while the model could be trained on monocular footage, it assumes a given camera calibration, which prevents arbitrary footage from the web from being used as training data.

In a concurrent study, Vijayanarasimhan et al.~addressed this shortcoming by explicitly modeling scene geometry \cite{vijayanarasimhan_sfm-net:_2017}. Inspired by geometrically-constrained Simultaneous Localization and Mapping (SLAM) models and Godard's work on left-right consistency, they proposed a model capable of detecting both ego-motion and object motion---as well as depth and object segmentation---from uncalibrated monocular images. Building upon those previous works, Mahjourian et al.~proposed a completely unsupervised model with explicit geometric scene modeling \cite{mahjourian_unsupervised_2018}. Their model introduced a new 3D loss function and added a new principled mask for handling unexplainable input.

In a second concurrent study, Godard et al.~ build on the concepts of \cite{vijayanarasimhan_sfm-net:_2017}.  They use a new appearance matching loss that better handles objects that are occluded curing the motion between frames, a new auto-masking approach to ignore portions of the image that aren't moving relative to the camera, and an improved multi-scale loss that reduces depth artifacts. The auto-masking is done by ignoring pixels where the reprojection decreases the similarity, and improves the handling of other moving objects. This notably decreases holes of large depth in roads, which can be seen later in this paper.

A recent study by van Dijk and de Croon \cite{dijk2019neural} explores what cues neural networks for monocular depth estimation use to predict depth maps.  Their study of four different networks found that the visual position of objects was the strongest cue for depth, rather than object size.  This property was found to hold regardless of the network architecture or training method (supervised or unsupervised).

\begin{figure*}[t]
    \begin{subfigure}[t]{0.7\linewidth}
        \centering
            \includegraphics[width=1\linewidth]{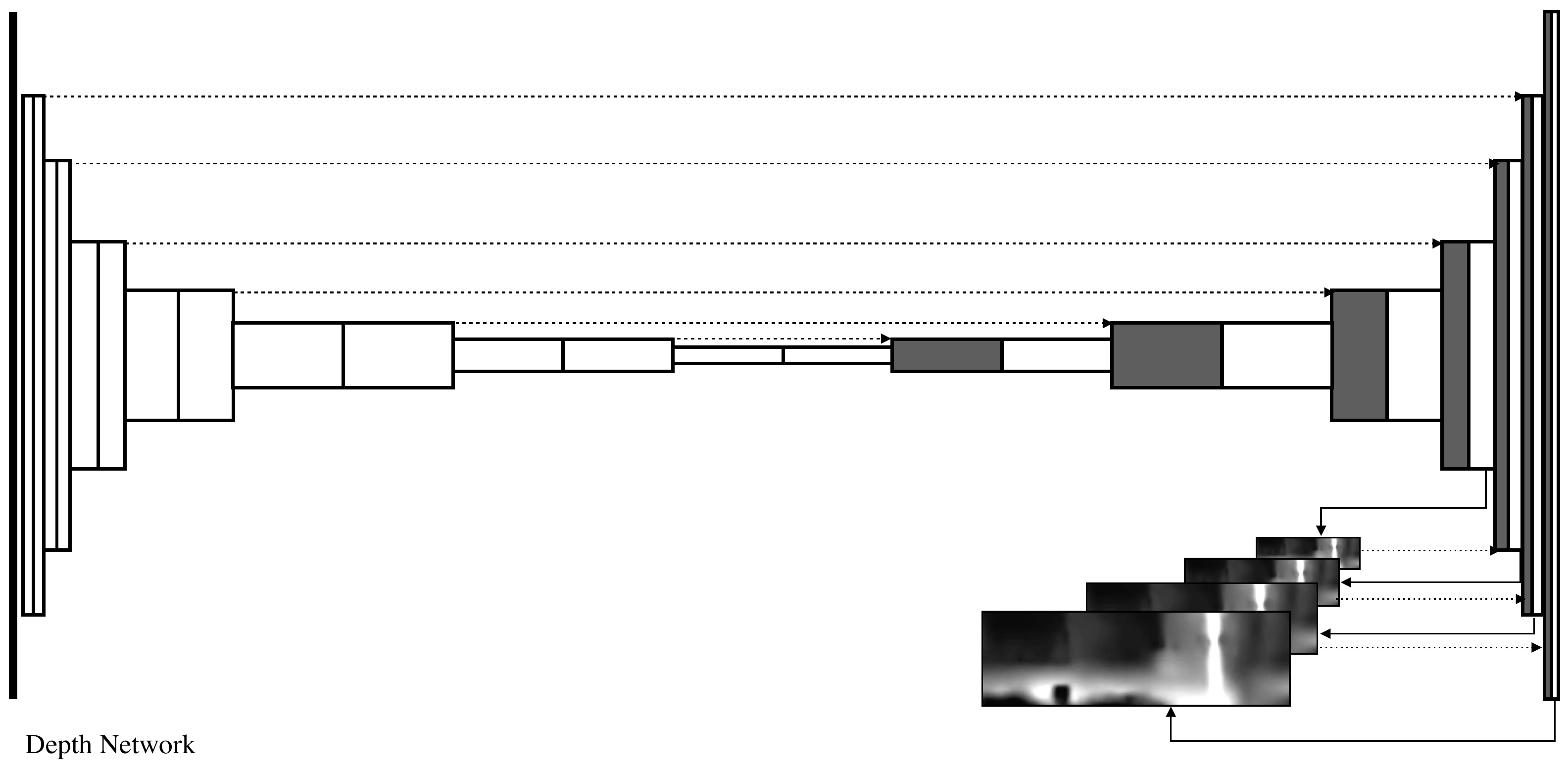}
        \caption{The depth network consists of a seven-layer encoder followed by a seven-layer decoder with a skip-layer architecture. The network returns the multi-scale disparity predictions, which can then be converted to depth predictions.}
        \label{fig:depth_network}
    \end{subfigure} \hfill
    \begin{subfigure}[t]{0.25\linewidth}
        \centering
            \includegraphics[width=1\linewidth]{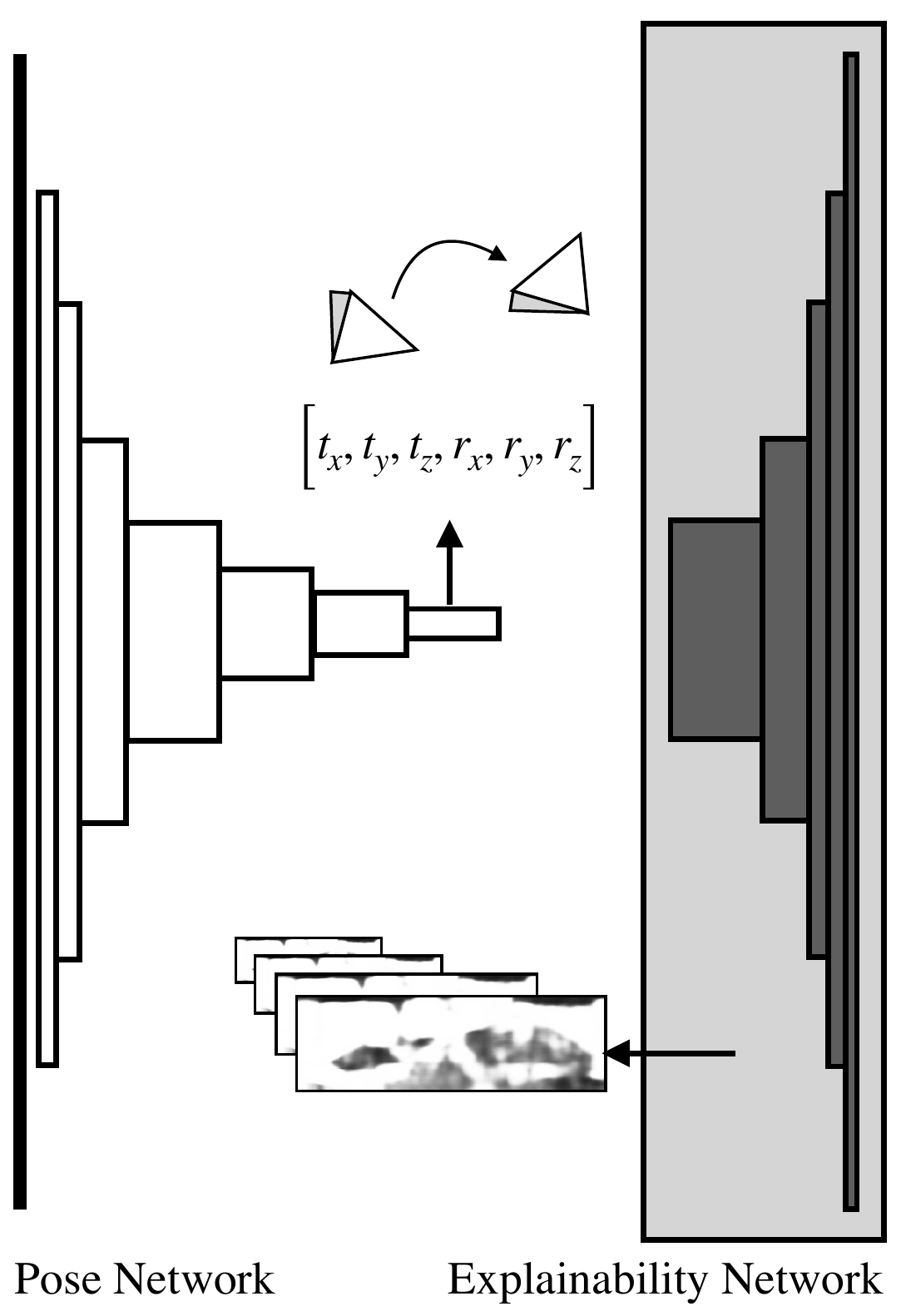}
        \caption{The pose network consists of a five-layer encoder followed by two pose layers and returns the pose as a six-element vector.}
        \label{fig:pose_network}
    \end{subfigure}
    \caption{The SfMLearner model \cite{zhou_unsupervised_2017} consists of two jointly-trained CNN stacks. The left diagram shows the depth CNN, and the right diagram shows the pose CNN.}
    \label{fig:proposed_model_layers}
\end{figure*}

\subsection{Beyond Pinhole Projection}

All of the models previously discussed take pinhole images as input. However, pinhole images have a serious disadvantage: objects can move out of the field of view.
Many applications, such as robotic navigation or virtual reality, benefit from the 360$^\circ$ field of view.
Previous research tackled this omnidirectional SfM problem using direct methods with some success \cite{huang_6-dof_2017},
but comparatively little research has been done with 360$^\circ$ imagery and CNNs.

Spherical input is particularly challenging for CNNs because spheres cannot be perfectly represented by a rectangular grid. In equirectangular projection, a common spherical projection method, this results in significant distortion in the polar regions of the image that propagate error through the convolutional layers. The traditional solution to this has been to use additional parameters and data augmentation to correct the distortions \cite{kumar_near-eld_2018}, but recently, researchers have presented several CNNs designed for direct spherical input.
Several of these approaches aim for full rotational invariance by using signal processing techniques to model spherical convolutional layers  \cite{cohen_spherical_2018,esteves_learning_2018,weiler_learning_2018}. While this approach is effective, it is also expensive; these models are severely limited in their input data size, making them impractical for most problems.

As full rotational invariance is often not required, several researchers have suggested a lighter-weight alternative: replacing normal convolutional filters with distortion-aware filters \cite{coors_spherenet:_2018,tateno_distortion-aware_2018,zhao_distortion-aware_2018}. In this approach, the standard rectangular CNN filter is replaced by a filter that samples points based on the image distortion, correcting the polar distortion effects. These models can be trained with one projection model and tested with another, allowing them to utilize the large body of pinhole datasets.
There were several other approaches aside from full rotational invariance and distortion-aware filters, including graph CNNs \cite{khasanova_graph-based_2017}, style transfer \cite{de_la_garanderie_eliminating_2018}, and increased filter sizes in the polar regions \cite{su_learning_2017}.

To avoid the distortions in the equirectangular projection, Cheng et al.~\cite{Cheng_2018_CVPR} and Wang et al.~\cite{wang_learning_2018} use cube maps as input to CNNs for saliency prediction and unsupervised depth and ego-motion.  They run a network on each cube face separately and use padding and specialized loss functions to encourage coherence between the outputs.

Cylindrical projection is another way of achieving 360$^\circ$ views around a given axis. Unlike spherical panoramas, cylindrical panoramas do not capture the full 3D space. However, they have a major benefit: they can be mapped exactly to a single plane, removing the issues of polar distortion and discontinuities.
Despite this benefit, little research has been done using deep networks with the cylindrical projection model \cite{esteves_polar_2018,salehinejad_cylindrical_2018}.
Furthermore, while some spherical CNN networks have made simplifications, including cropping of the polar regions \cite{coors_spherenet:_2018} and distortion-aware filters \cite{coors_spherenet:_2018}, to the best of our knowledge, no previous work has applied a cylindrical CNN to the structure-from-motion problem.
Our work introduces the first CNN model designed to predict depth and pose from cylindrical panoramic input.

\section{Methods}

In this work, we present an unsupervised convolutional model that jointly estimates the depth map from a single cylindrical panoramic image and ego-motion from a short image sequence.

\subsection{Model Architecture}

Our architecture is based on that of Zhou et. al \cite{zhou_unsupervised_2017}, an unsupervised model designed to predict depth and ego-motion in monocular pinhole images. The architecture, illustrated in Figure \ref{fig:proposed_model_layers}, is a convolutional network consisting of two jointly-trained stacks:
(a) a depth network to estimate the depth map,
(b) a pose network mask to estimate the change in the pose in image sequences and handle unexplainable input.
The depth network follows the DispNet \cite{mayer_large_2016} skip-layer architecture, with seven contracting layers and seven expanding layers, outputting a multi-scale depth prediction. The pose network (PoseExpNet) consists of five contracting convolutional layers and three pose layers, outputting the predicted translation and rotation between the source and target views.

\begin{figure}[h] 
    \centering
    \includegraphics[width=0.4\linewidth]{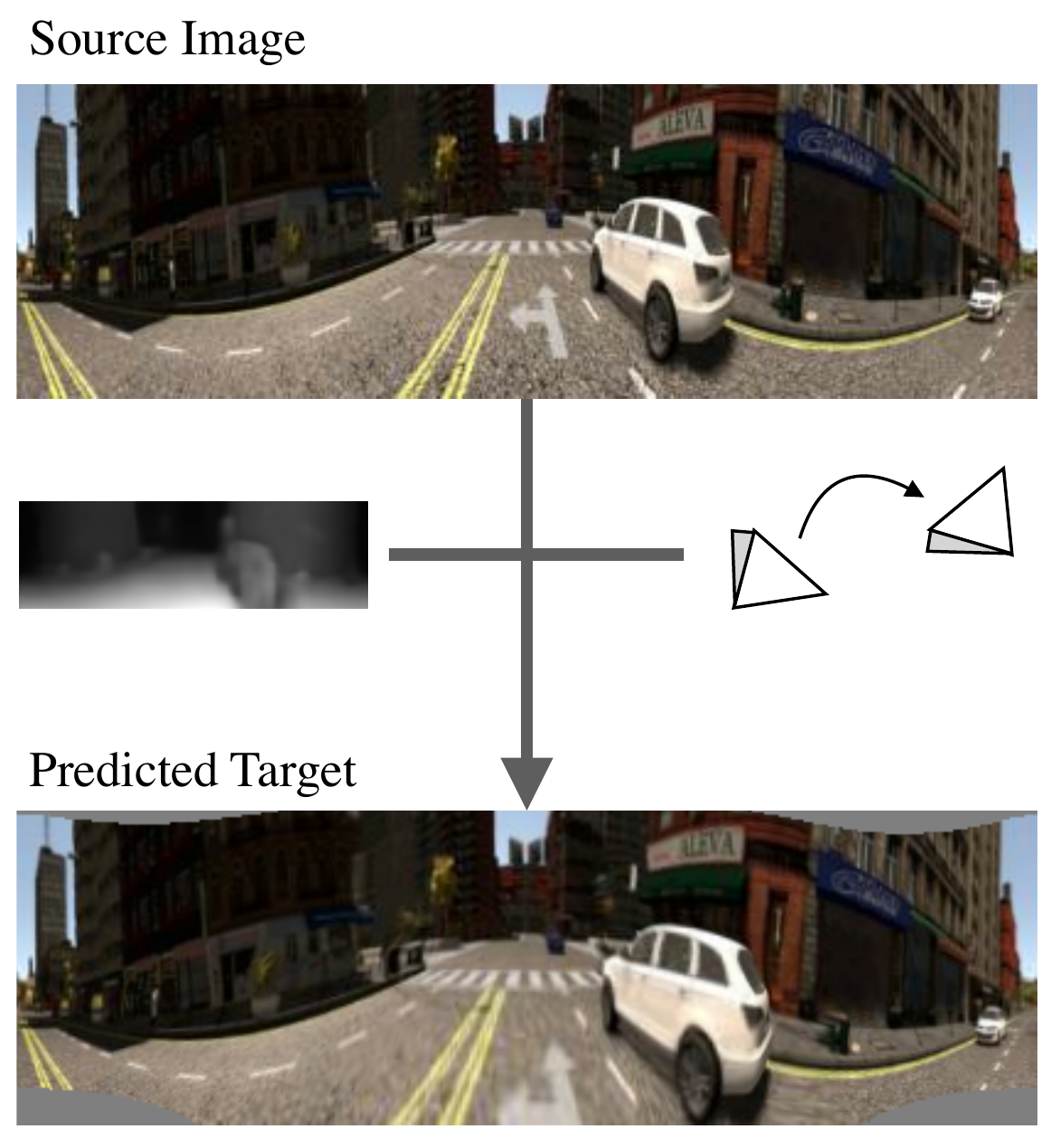}
    \caption{We use panoramic view synthesis as a supervisor: the source panorama, depth, and pose transformation are used to synthesize a target view, and the loss is computed as the difference between the actual and synthesized views.  As the synthesized view improves, the depth and pose predictions improve.}
    \label{fig:proj_warp}
\end{figure}

The learner uses \textit{view synthesis}---the prediction of a target frame given source frames---as a self-supervisory signal (Figure \ref{fig:proj_warp}).  The network takes one target frame and several neighboring source frames as input. At each training step, the joint DispNet and PoseExpNet stacks predict (a) the depth of the target frame and (b) the pose of each source frame in the target frame's coordinate system.  These predictions are then applied to the source images to synthesize the target image through {\it projective inverse warping}.

Let $\mathcal{D}$ be the predicted depth map for the target image.  The corresponding 3D point $X^{i,j}_{\text{target}}$ for pixel $i,j$ in the target image is found by unprojection according to the predicted depth value at $\mathcal{D}^{i,j}$.
\begin{equation}
X^{i,j}_{\text{target}} = \text{unproject}(i,j,\mathcal{D}^{i,j})
\end{equation}
This 3D point is then transformed into the source image's coordinate frame according to $\mathcal{P}_{\text{source}}$ the predicted pose of the source image.
\begin{equation}
X^{i,j}_{\text{source}} = \text{transform}(\mathcal{P}_{\text{source}},X^{i,j}_{\text{target}})
\end{equation}
Finally, the source image is sampled using bilinear interpolation at the projection of $X^{i,j}_{\text{source}}$.
\begin{equation}
\mathcal{I}^{i,j}_\text{proj} = \text{sample}(\mathcal{I}_{\text{source}},\text{project}(X^{i,j}_{\text{source}}))
\end{equation}
The learner then tries to minimize the photometric error between the $\mathcal{I}_{\text{target}}$ and $\mathcal{I}_{\text{proj}}$, the source image warped into the target image's coordinate frame. See Figure \ref{fig:cylindrical_proj} for an illustration.
The learner reduces the photometric error by improving the depth and pose predictions \cite{zhou_unsupervised_2017}, allowing the network to learn scene structure from monocular images without labeled depth maps.

For this model, we use a three-part objective function. The main component is the \textit{photometric loss} ($\mathcal{L}_\text{pixel}$), which minimizes the difference between synthesized views and the target view. This is regularized by the \textit{smooth loss} ($\mathcal{L}_\text{smooth}$), which minimizes the second derivatives with respect to the depth. If $\lambda_s$ and $\lambda_e$ represent the smooth weights, the total loss can be written as follows:

\begin{equation}
    \mathcal{L} = \sum_\text{scales} \left( \sum_\text{sources}\mathcal{L}_\text{pixel} + \lambda_s \mathcal{L}_\text{smooth} + \sum_\text{sources}\lambda_e \mathcal{L}_\text{exp} \right)
    \label{eq:total_loss}
\end{equation}

If $\mathcal{I}$ is an RGB image, $\mathcal{D}$ is the depth prediction, the three loss components at pixel $i,j$ at each scale can be written as follows:

\begin{equation}
    \mathcal{L}_\text{pixel}^{i,j} =
    \sum_{i,j} \mathcal{E}^{i,j} \left| \mathcal{I}_\text{proj}^{i,j} - \mathcal{I}_\text{target}^{i,j} \right|
    \label{eq:pixel_loss}
\end{equation}
\begin{equation}
    \mathcal{L}_\text{smooth}^{i,j} =
        \left\lvert\frac{\delta^2 \mathcal{D}^{i,j}}{\delta x^2}\right\rvert +
        \left\lvert\frac{\delta^2 \mathcal{D}^{i,j}}{\delta x \delta y}\right\rvert +
        \left\lvert\frac{\delta^2 \mathcal{D}^{i,j}}{\delta y \delta x}\right\rvert +
        \left\lvert\frac{\delta^2 \mathcal{D}^{i,j}}{\delta y^2}\right\rvert
    \label{eq:smooth_loss}
\end{equation}
\begin{equation}
\mathcal{L}_\text{exp}^{i,j} = \sum\text{softmax}\left(\mathcal{E}^{i,j}\right)
\end{equation}

We also experimented with the image-aware depth smoothness loss term introduced by Wang et al. \cite{wang_learning_2018}:
\begin{equation}
\mathcal{L}_\text{smooth}^{i,j} = e^{-\nabla^2 \mathcal{I}^{i,j}} (
| \frac{\partial \mathcal{D}^{i,j}}{\partial x^2} | +
| \frac{\partial \mathcal{D}^{i,j}}{\partial x \partial y} | +
| \frac{\partial \mathcal{D}^{i,j}}{\partial y^2} |)
\end{equation}
The advantage of this smoothing term is that it reduces the depth smoothing effect at image edges, whereas the previous smoothing term (\autoref{eq:smooth_loss}) smooths the entire depth map uniformly without respect for image edges.

Two major modifications were required to allow for cylindrical input: 1) the view synthesis functions were modified to account for cylindrical projection, and 2) the convolutional layers, resampling functions, and loss were modified to preserve horizontal wrapping.

\subsubsection{Camera Projection and Cylindrical Panoramas}

The view synthesis function introduced by Zhou et al. \cite{zhou_unsupervised_2017} works by warping the source images onto the target image's coordinate frame using the predicted target inverse depth map and the source image relative pose.  To adapt this process to work with cylindrical input, we modified the mapping functions between the pixel, camera, and world coordinate frames.

\begin{figure}[t] 
    \centering
    \includegraphics[width=0.4\linewidth]{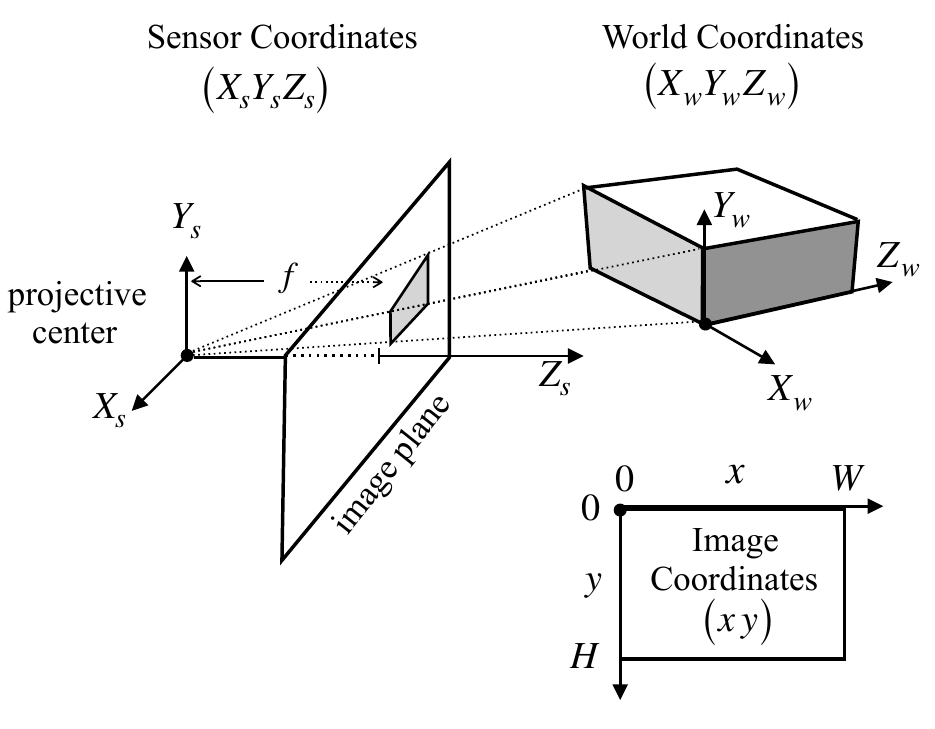}
    \caption{Pinhole projection model. The 3D world (on the world coordinate system) is projected on a flat image plane; the image plane is a focal length $f$ away from the projective center along the $Z_s$ axis (on the sensor coordinate system). The result is a $W \times H$ rectangular image.}
    \label{fig:pinhole_proj}
\end{figure}

Most structure-from-motion systems expect \textit{pinhole projection} images as input. Pinhole projection images project a 3D scene from the world coordinate system onto a flat image plane; this process can be described by the focal length $f$, principle point $c$, and the image height $H$ and width $W$, as shown in Figure \ref{fig:pinhole_proj}.

\begin{figure}[t] 
    \centering
    \includegraphics[width=0.45\linewidth]{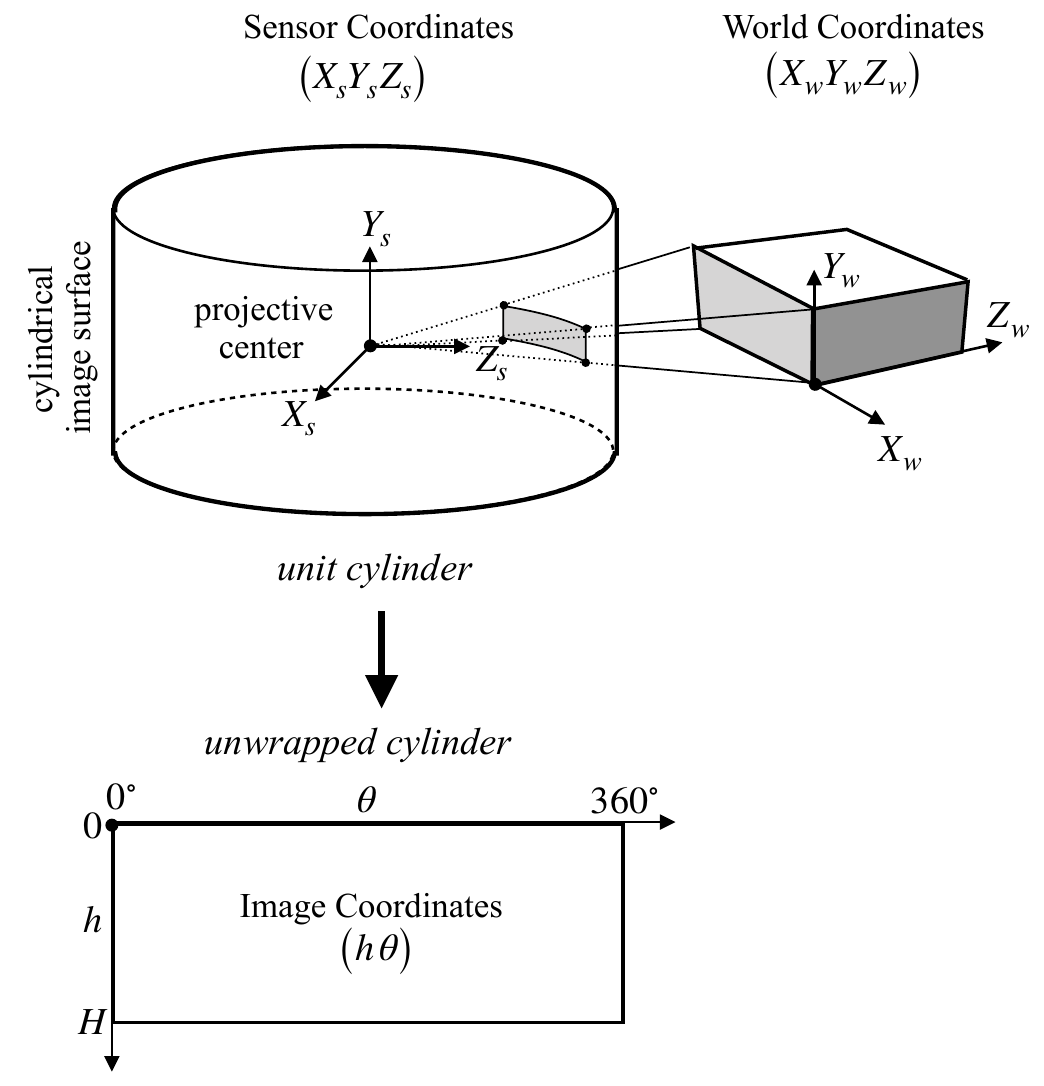}
    \caption{Cylindrical projection model. In contrast with pinhole projection, this projects an image onto a curved cylindrical surface. The final result is a rectangular image with height $H$ and a width representing the full $360^\circ$.}
    \label{fig:cylindrical_proj}
\end{figure}

In contrast, \textit{cylindrical projection} projects the 3D world onto a curved cylindrical surface, as seen in Figure \ref{fig:cylindrical_proj}. The goal of this process is to take a 3D point in the world coordinate system and project it onto a rectangular cylindrical panorama. This requires projecting the 3D point onto the cylindrical image surface and converting the image surface into a Cartesian coordinate system.

The transformation between the sensor and pixel coordinate systems can be described by the following equations. A 3D point $P = \left(x_s, y_s, z_s\right)$ in the sensor coordinate system projects to a 2D point $Q = \left(\theta, h\right)$ on the unit cylinder around the origin according to the following formula:

\begin{equation}
\begin{bmatrix}
    \theta \\
    h
\end{bmatrix} =
\begin{bmatrix}
    \arctan\left(\frac{x_s}{z_s}\right) \\
    \frac{y_s}{\sqrt{x_s^2+z_s^2}}
\end{bmatrix}
\end{equation}

The inverse projection from the unit cylinder to a 3D point in the sensor coordinate system is as follows:
\begin{equation}
    \label{eq:inverseproj}
    \begin{bmatrix}
        x_s \\
        y_s \\
        z_s
    \end{bmatrix} =
    \begin{bmatrix}
        d \sin{\theta} \\
        d h \\
        d \cos{\theta}
    \end{bmatrix}
\end{equation}
where $d$ is the depth of the point.

For our experiments, we modified the Tensorflow implementation of SfMLearner by Zhou et al.~\cite{zhou_unsupervised_2017} to use these cylindrical un-projection and projection functions.

\subsubsection{Horizontal Wrapping}

In order to extend the model for cylindrical panoramic images, we  modified the convolutional layers, smooth loss function, and 2D projection to account for horizontal wrapping.

\begin{figure}[b]
    \centering
    \includegraphics[width=0.4\linewidth]{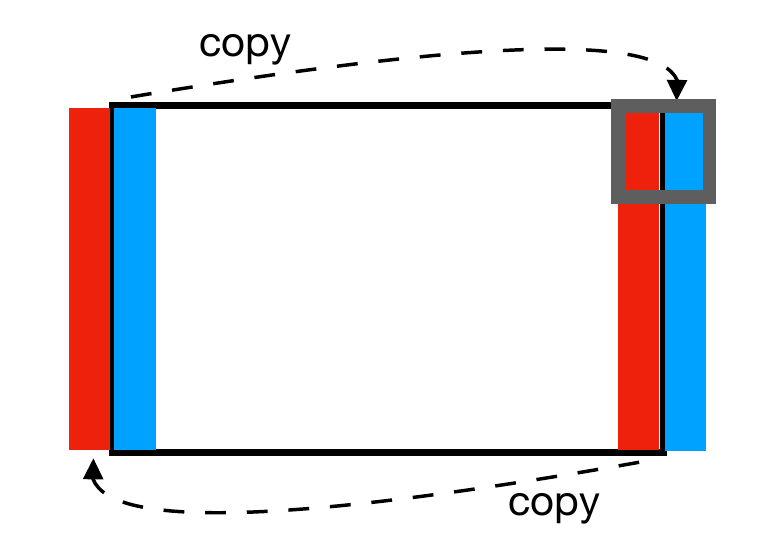}
    \caption{An example of a horizontally wrapping convolutional layer.  The wrapping is achieved by copying the left-most columns to the right side and vice-versa.}
    \label{fig:horiz_wrapping}
\end{figure}

Unlike pinhole images, cylindrical images wrap horizontally.   For a convolutional layer to work with cylindrical input, it must preserve this horizontal wrapping property, rather than using zero-padding as is typical.  Horizontal wrapping can be done by padding the right side of the tensor with columns from the left and vice-versa, as depicted in Figure \ref{fig:horiz_wrapping}.

\begin{figure*}[t]
    \centering
    \includegraphics{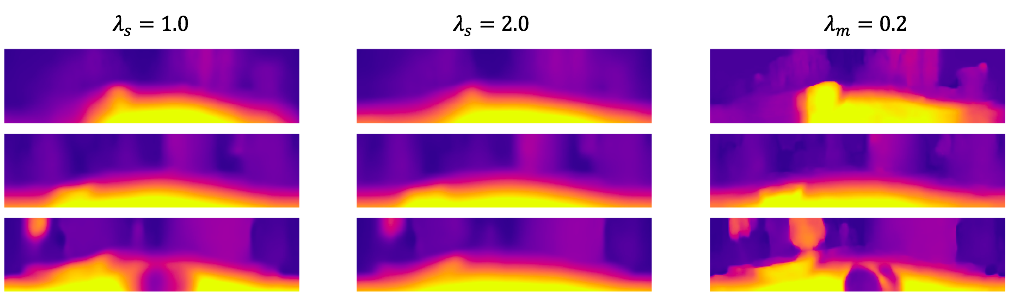}
    \caption{A comparison of different loss functions. The first and second columns represent second-order depth gradient loss at different weights: $\lambda_s=1.0$ and $\lambda_s=2.0$. The rightmost column shows an image-aware first-order gradient loss with a weight $\lambda_m=0.2$.}
    \label{fig:loss_comparison}
\end{figure*}

\begin{table*}[b]
    \centering
    \small\addtolength{\tabcolsep}{-2pt}
    \begin{tabular}{ | l || c | c | c | c || c | c | c | }
        \multicolumn{1}{c}{} & \multicolumn{4}{c}{Disparity (lower is better)} & \multicolumn{3}{c}{Accuracy (higher is better)} \\
        \hline
        ~ & Abs Rel & Sq Rel & RMSE & RMSE log & $\delta < 1.25$ & $\delta < 1.25^2$ & $\delta < 1.25^3$ \\ \hline
        No wrapping & 0.301 & 2.543 & 6.220 & 0.395 & 0.668 & 0.828 & 0.901 \\
        Wrapping & \textbf{0.299} & \textbf{2.514} & \textbf{6.037} & \textbf{0.387} & \textbf{0.675} & \textbf{0.835} & \textbf{0.906} \\
        \hline
        \end{tabular}
    \caption{Depth accuracy results on SYNTHIA when training and testing our network architecture with and without horizontal wrapping.  The model trained with horizontal wrapping is better on all metrics.}
    \label{table:ablation}
\end{table*}

We added wrap padding to all convolutions in the network architecture.  We also added wrap padding to the smooth gradient loss computation and the bilinear sampler used for view synthesis.

\section{Experiments}

Several experiments were performed to evaluate the performance of our proposed method.
We performed an ablation study on two synthetic datasets, comparing the model with and without horizontal wrapping.  We also evaluated the accuracy of pose estimation with different input fields of view.  Finally, to demonstrate the model's effectiveness on real-world footage, we trained and tested the network on a new panoramic video dataset and qualitatively evaluated the results.

CylindricalSfMLearner was implemented in the open-source machine learning library Tensorflow \cite{tensorflow2015-whitepaper}.
Our code is publicly available on GitHub at \href{https://github.com/jonathanventura/cylindricalsfmlearner}{https://github.com/jonathanventura/cylindricalsfmlearner}.

\subsection{Monocular Depth Estimation}

While there are many standard datasets such as KITTI \cite{geiger_are_2012} and CityScapes \cite{cordts_cityscapes_2016} that provide standard field-of-view video with registered ground truth depth, we were unable to find a similar dataset containing panoramic video and associated ground truth with suitable characteristics.  For this reason, we performed a quantitative evaluation of our work on synthetic data.

SYNTHIA-Seqs \cite{ros_synthia_2016} is a synthetic dataset of driving data designed to mimic the properties of popular front-view datasets like KITTI \cite{geiger_are_2012} and CityScapes \cite{cordts_cityscapes_2016}.
For this experiment, we used sequences 02 and 05 (NYC-like city driving) in the spring-, summer-, and fall-like conditions captured from the left and right stereo cameras.

To prepare the data for training, we first stitched the four perspective views into a single 360$^\circ$ cylindrical panorama.
Next, we identified static frames using the global pose ground truth; these frames were excluded from the final formatted dataset using the same technique as SfMLearner \cite{zhou_unsupervised_2017}. The panoramas were then resized to $512 \times 128$ and concatenated to form three-frame sequences.
After processing, this dataset contained 4,398 panoramic views. This was split into train, test, and validation subsets comprised of approximately 80\%, 10\%, and 10\% of the data, respectively.

We experimented with various settings of the smoothing term and also alternative smoothing terms such as the image-aware first-order gradient loss from Wang et al.\cite{wang_learning_2018}.  A comparison is shown in Figure \ref{fig:loss_comparison}.  We found that while increasing the smoothness reduces the detail in the depth prediction, it tended to increase the depth accuracy.  For the remaining experiments on SYNTHIA, we chose a smoothness term of $\lambda_s = 2.0$ and $\lambda_e = 0$ and trained all models for 60,000 steps.

We compared the model with and without horizontal wrapping to determine whether the wrapping property is important for prediction accuracy.
We used a set of disparity and accuracy metrics common in SfM research that each capture a different aspect of the prediction error; for more details, please refer to \cite{eigen_depth_2014}.

Table \ref{table:ablation} shows the results of our experiment.   The model trained with horizontal wrapping in the convolutional layers outperformed the model without wrapping on all metrics.

\subsection{Pose Estimation}

We hypothesized that one benefit of end-to-end training of depth and ego-motion estimation with panoramic video would be an improvement in pose accuracy.  We reasoned that the larger field-of-view (FOV) would provide more rays constraining the camera pose and thus improve pose estimation accuracy.  To test this hypothesis we trained different models on crops of the full cylindrical panorama to simulate cameras of different FOVs.  We did not use horizontal wrapping on the cropped images.

We trained the models on data from SYNTHIA sequences 02 and 05 and tested on unseen data from sequences 02, 04, and 05. This split was selected due to the limited number of visually similar SYNTHIA sequences. Sequences 02 and 05 were partitioned randomly into training and testing sets, and no frame seen during training as either source or target frame was used in the testing set.

Table \ref{table:pose} reports the mean average trajectory error (ATE) \cite{mur-artal_orb-slam:_2015} achieved by each model on SYNTHIA sequences 02, 04, and 05.
Sequence 04 is visually dissimilar from the training data, which explains why the ATE is higher on sequence 04.  In general, the higher FOV models have lower ATE.

\begin{table}[t]
    \centering
    \small
    \begin{tabular}{ | r | r | l || r | r | }
    \hline
    Seq. & FOV (deg) & Wrap & ATE Mean & ATE Std. Dev. \\
    \hline
    02 & 360 & Yes & \textbf{0.0139} & \textbf{0.0210} \\
    02 & 360 & No & 0.0143 & 0.0211 \\
    02 & 270 & No & 0.0150 & 0.0230 \\
    02 & 180 & No & 0.0203 & 0.0296 \\
    02 & 100 & No & 0.0324 & 0.0498 \\
    \hline
    05 & 360 & Yes & 0.0138 & 0.0195 \\
    05 & 360 & No & 0.0132 & \textbf{0.0189} \\
    05 & 270 & No & \textbf{0.0131} & 0.0191 \\
    05 & 180 & No & 0.0167 & 0.0224 \\
    05 & 100 & No & 0.0258 & 0.0383 \\
    \hline
    04 & 360 & Yes & \textbf{0.0342} & 0.0351 \\
    04 & 360 & No & 0.0345 & \textbf{0.0322} \\
    04 & 270 & No & 0.0355 & 0.0348 \\
    04 & 180 & No & 0.0384 & 0.0372 \\
    04 & 100 & No & 0.0371 & 0.0359 \\
    \hline
\end{tabular}
\caption{Evaluation of the effect of input field-of-view on average trajectory error (ATE).  In general, a wider field-of-view leads to higher pose accuracy (lower mean ATE).  The training data contained images from Sequences 02 and 05 but not Sequence 04.}
\label{table:pose}
\end{table}

\subsection{CARLA synthetic dataset}

SYNTHIA is a very useful dataset for testing cylindrical depth perception, however, it is limited to one vehicle trajectory per city -- the different sequences in each city have different illumination but the same trajectory.  Each city also varies widely in appearance, making it difficult for the neural network to generalize to unseen data.  We wanted to be able to test on lots of data from relatively similar cities, in part so that we could test on a never before seen city that is still relatively similar to the training data.

\begin{figure}[hbt!]
    \centering
    \includegraphics{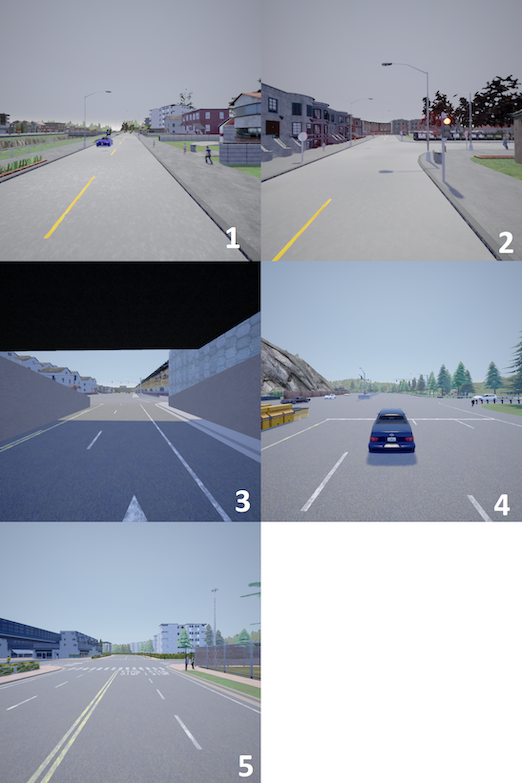}
    \caption{Images from each of CARLA's towns.  In left to right row major order: 1, 2, 3, 4, 5}
    \label{fig:carla_towns}
\end{figure}

To this end, we used the CARLA simulator \cite{Dosovitskiy17} to create a new synthetic dataset.  Using the simulator, we could generate many different trajectories in each city, and had more options for different cities with similar visual appearance. CARLA also made it easy for us to include many dynamic objects such as other cars, bycicles, motorcycles, and pedestrians.  Traffic rules are also followed, including interactions with other cars or pedestrians.  There are five cities, varying from a small mountainous town to a large city.  CARLA provides varied driving environments, including freeways and tunnels.  We used cities 1,3,4, and 5 for training data and tested on city 2.  The weather and illumination conditions were set to clear weather at noon.  Images of each city are shown in \autoref{fig:carla_towns}.

Because CARLA does not support panorama output, we rendered six-sided cube maps and stitched them into panoramas afterward.  We rendered 100 degree FOV $768 \times 768$ images of depth and RGB for each of the six cube sides, and recorded pose data.  Data was recorded at 5 fps (the same as SYNTHIA).  Each drive is $1,000$ frames long, and we recorded 5 drives for each city.  We stitched the  images (RGB and depth) into $2048 \times 1024$ cylindrical panoramas.

\subsubsection{Training}

Evaluation was done using the same method as used on SYNTHIA, with images processed at their full size of $2048 \times 1024$.  Ground truth pose was used to remove static frames.  After this was done, we had $11,604$ 3-frame sequences, more than twice as much as SYNTHIA.  The models were trained using image-aware smooth loss ($\lambda_m = 0.2$).

\begin{figure}[hbt!]
    \centering
    \includegraphics{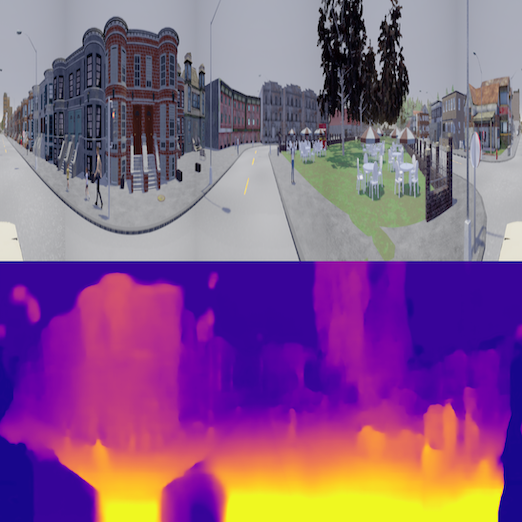}
    \caption{A sample of our CARLA dataset's cylindrical color images, and the predicted depth.}
    \label{fig:carla_sample}
\end{figure}

A sample cylindrical RGB image and the predicted depth map is shown in \autoref{fig:carla_sample}.  The visual quality of the simulation is somewhat lower than SYNTHIA's, but it includes more dynamic objects (cars, motorcycles, bikes, and pedestrians).

\subsubsection{Monocular Depth Estimation}

The results of our depth evaluation are shown in Table \ref{table:carla_eval}.  With wrapping is notably superior.  The metrics are bit worse than SYNTHIA's.  Some likely reasons for this are less hyperprameter optimization, lower texture quality, more dynamic objects, and using a never seen city as the test set.

 A sample depth prediction is shown in \autoref{fig:carla_sample}.  Qualitatively, the models predictions are quite good.  Dynamically moving objects (other cars, pedestrians) are handled quite well.
 Fences and other fine structures appear in the predicted depth maps but are not always perfectly estimated.
One issue we noticed in predictions on this dataset and the other dataset is the presence of relatively holes of large depth in the street regions.  These can be seen in the bottom-right and especially the right hand side of \autoref{fig:carla_sample}.

\begin{table*}[h!]
    \centering
    \small\addtolength{\tabcolsep}{-2pt}
    \begin{tabular}{ | l || c | c | c | c || c | c | c | }
        \multicolumn{1}{c}{} & \multicolumn{4}{c}{Disparity (lower is better)} & \multicolumn{3}{c}{Accuracy (higher is better)} \\
        \hline
        ~ & Abs Rel & Sq Rel & RMSE & RMSE log & $\delta < 1.25$ & $\delta < 1.25^2$ & $\delta < 1.25^3$ \\
        \hline
        No wrapping & 0.550 & 16.266 & 12.868 & 0.560 & 0.475 & 0.745 & 0.856 \\
        Wrapping & \textbf{0.497} & \textbf{14.212} & \textbf{12.032} & \textbf{0.528} & \textbf{0.565} & \textbf{0.781} & \textbf{0.867} \\
        \hline
    \end{tabular}
    \caption{Depth accuracy results when training and testing our network architecture with and without horizontal wrapping, on the CARLA dataset.  The model trained with horizontal wrapping is better on all metrics.}
    \label{table:carla_eval}
\end{table*}

\subsection{Real data: Headcam Dataset}

\begin{figure*}[ht]
    \centering
    \includegraphics{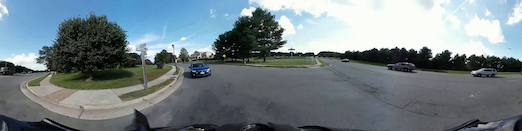}\\
    \includegraphics{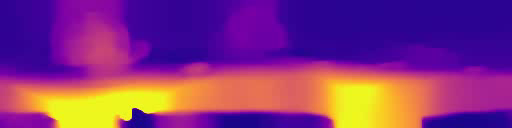}\\
    \includegraphics{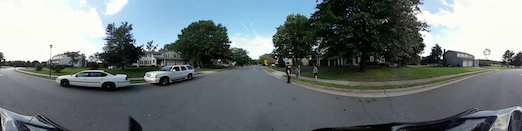}\\
    \includegraphics{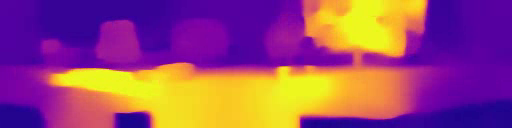}\\
    \caption{Input images and depth predictions for the Headcam dataset when trained on $512 \times 128$ images.}
    \label{fig:headcam_small}
\end{figure*}

\begin{figure*}[ht]
    \centering
    \includegraphics{000001_large_input.png}\\
    \includegraphics{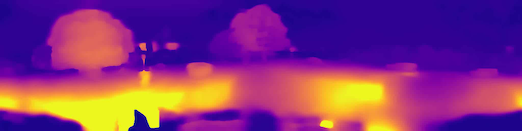}\\
    \includegraphics{000100_large_input.png}\\
    \includegraphics{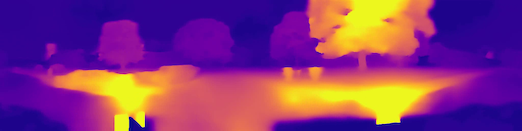}
    \caption{Input images and depth predictions for the Headcam dataset when trained on on $2048 \times 512$ images.  Training on larger images results in better estimation of shapes and fine structures but also introduces some artifacts.}
    \label{fig:headcam_large}
\end{figure*}

While the SYNTHIA-Seqs and CARLA datasets allowed us to explore the effect of cylindrical projection and increasing field of views, its major limitation is that it is synthetic. In order to determine if cylindrical projection is successful with real-world input, we trained and evaluated a model on our own panoramic dataset.

We assembled a panoramic video dataset which we call Headcam.  It was collected by affixing a consumer-grade panoramic camera, the 2016 Samsung Gear 360, to a bicycle helmet and biking around neighborhoods in Northern Virginia.  This dataset includes about two hours of footage collected over  three days and has been released publicly on Zenodo under a Creative Commons Attribution license
\cite{zenodo}.

This footage was first stitched into equirectangular panoramic video using the Samsung Gear 360 software. It was then broken into frames at 5fps, warped into a cylindrical projection model, resized to $2048 \times 512$, and formatted into three-frame sequences.
The final formatted dataset contains 27,538 frames.
Training was conducted on 90\% of the data, and qualitative testing was done on the remaining 10\%.

This model was trained with the same configuration as SYNTHIA except for the smooth loss term. This model was trained using image-aware smooth loss ($\lambda_m = 0.2$) \cite{wang_learning_2018}as opposed to the second-order gradient loss. This change helped to improve the definition of predicted object shapes such as trees and cars.

We compared the result of training and testing with small ($512\times128$) versus large ($2048\times512$) panoramas with the Headcam dataset.  \autoref{fig:headcam_small} shows the result with small input images and \autoref{fig:headcam_large} shows the result with large images.  With large images, the shapes of the trees and cars are noticably improved.  Additionally, fine structures such as the street signs, which are omitted from the small depth maps, are present in the large depth maps.  However, the large depth maps sometimes have more artifacts around fine structures.

\begin{figure}[t]
    \centering
    \includegraphics{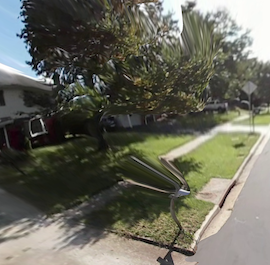}
    \includegraphics{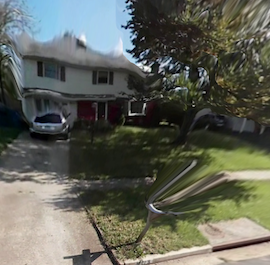}
    \includegraphics{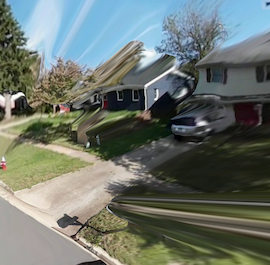}\\
    \includegraphics{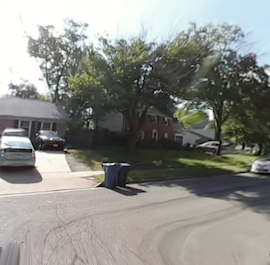}
    \includegraphics{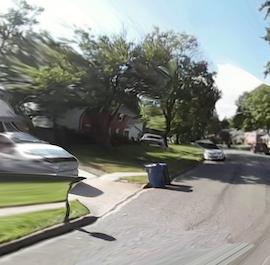}
    \includegraphics{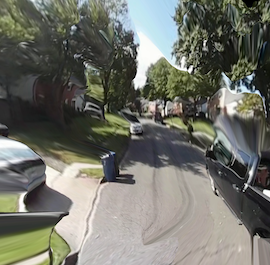}\\
    \caption{Examples of new view synthesis using the estimated depth map to create a textured mesh.}
    \label{fig:viewsynth}
\end{figure}

\begin{figure}[t]
    \centering
    \includegraphics{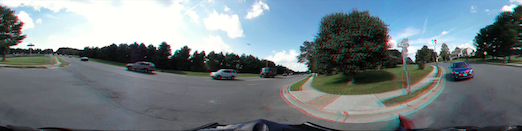}
    \includegraphics{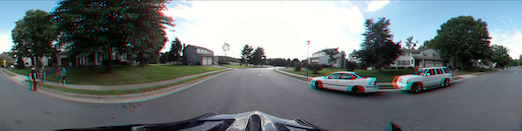}
    \includegraphics{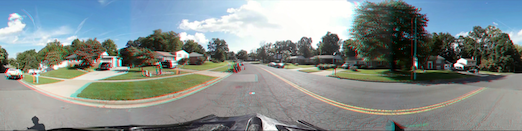}
    \includegraphics{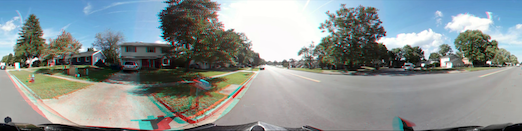}
    \includegraphics{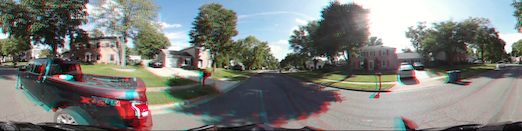}
    \caption{Generated stereo panoramas shown as red–blue anaglyphs.}
    \label{fig:stereopano}
\end{figure}

Qualitatively, the model generates visually reasonable predictions with well-defined object boundaries. However, the model makes several common errors. Most notably, due to its constant presence frame-to-frame, the model predicts that the helmet at the bottom of the image has a large depth. Future work might explore ways to mitigate this through masking or other techniques. We also noticed occasional holes in the depth prediction in the road, similar to the predictions on CARLA and on SYNTHIA when using a low weight on the smoothing term.

\section{Applications: New View Synthesis and Stereo Panorama Conversion}

We applied this network to a practical problem by using the predicted depth map to convert the input panorama into a stereo panorama \cite{peleg1999stereo}.  A stereo panorama consists of a left-eye and right-eye panorama so that, when viewed in a VR reality headset, the viewer perceives depth in the scene from parallax cues.  Normally, to produce a stereo panorama once must either spin a camera in a circle \cite{Megastereo} or use a multi-camera rig \cite{Jump2016,facebook}.  With our approach, we can produce a stereo panorama from a single input image captured with a standard $360^\circ$ camera.

\subsection{Mesh Creation}

To create the stereo panorama from a single input panorama, we use the predicted depth map output by our trained network to create a dense mesh and then synthetically render the left- and right-eye panoramas.

At test time we use the trained depth network to predict the depth of each pixel in the input panorama.  We create a triangular mesh representing the cylindrical input panorama with a vertex on each pixel location and edges between neighboring vertices.  The depth of each vertex is scaled by the corresponding depth in the predicted depth map.  Formally, the position of each vertex is $(d \sin \theta, d h, d \cos \theta)$ as given in Equation \ref{eq:inverseproj}.  The color of each vertex is set to the color of the corresponding pixel in the input panorama.

\subsection{New View Synthesis}

Once we have the textured mesh, we can easily render new viewpoints using the OpenGL rasterizer, for example.  We render without shading or lighting enabled so that the original color of the input panorama is rendered without modification.  Examples of new view synthesis are shown in Figure \ref{fig:viewsynth}.

\subsection{Stereo Panorama Rendering}

To render the stereo panorama, we use mesh-based view synthesis to render the virtual views around a unit circle.  We rasterize the stereo panoramas column by column, moving the virtual camera to each corresponding point on the unit circle \cite{googleods}.  Example stereo panoramas are shown in Figure \ref{fig:stereopano} as red-blue anaglyphs.

The stereo panoramas are viewable in a motion-tracked VR headset such as the Oculus Go.  Example panoramas can be viewed in a WebVR-compatible browser at our project webpage\footnote{\url{https://jonathanventura.github.io/cylindricalsfmlearner/}}.

\section{Conclusion}

We introduce the first unsupervised model for learning depth and ego-motion directly from panoramic video input.  In contrast to previous work, we use the cylindrical panoramic projection which allows for direct application of existing convolutional neural network models to panoramic input with little modification.  Our evaluation on synthetic and real data shows that learning from cylindrical panoramic input is as effective as pinhole projection input in producing accurate and detailed depth predictions.  We demonstrated how our method can be used to synthesize stereo panoramas from monocular input to support virtual reality content capture with commonly available consumer hardware.

We also contribute a novel dataset of real panoramic video suitable for unsupervised learning of depth and ego-motion.  The dataset was captured on city streets from a helmet-mounted camera while riding a bicycle and thus contains a variety of motions, moving objects, and other challenging conditions, making it a difficult and interesting dataset for future research on learning panoramic structure-from-motion.

While cylindrical projection makes the application of convolutional neural networks to panoramic input simple, this representation has some limitations.  First, the top and bottom of the complete spherical field-of-view are not included in the cylindrical representation; however, in most applications, these regions are not important.  Second, our model does not achieve full rotation invariance like, for example, the Spherical CNN model of Cohen et al. \cite{cohen_spherical_2018}.  Our cylindrical CNN model is invariant to rotation about the vertical axis (yaw) but not rotation about the other two axes (pitch and roll).  However, pitch and roll rotations of the camera are relatively rare in street-level driving tasks as evaluated in this paper.

Future work could also include a comparison of our cylindrical model against the various spherical CNN and cube map CNN models that have been proposed, and experimenting with other state-of-the-art depth prediction architectures \cite{yin_geonet:_2018,xu_structured_2018} adapted for cylindrical panoramic input.

\section*{Acknowledgments}

This material is based upon work supported by the National Science Foundation under Grant Nos. 1659788 and 1464420.

\bibliographystyle{ieeetr}
\bibliography{ms}

\end{document}